\documentclass{article}
\usepackage{amsmath,epsfig}
\usepackage{amsfonts,bm,multirow}
\usepackage[preprint]{spconfa4}

\copyrightnotice{978-1-7281-1331-9/20/\$31.00~\copyright 2020 IEEE}

\let\OLDthebibliography\thebibliography
\renewcommand\thebibliography[1]{
  \OLDthebibliography{#1}
  \setlength{\parskip}{0pt}
  \setlength{\itemsep}{0pt plus 0.3ex}
}

\begin{document}\sloppy

\def\x{{\mathbf x}}
\def\L{{\cal L}}

\title{Weakly Supervised Video Anomaly Detection via Center-guided Discriminative Learning}
%
\name{Boyang~Wan, Yuming~Fang, Xue~Xia, Jiajie~Mei}
\address{School of Information Technology, Jiangxi University of Finance and Economics, Nanchang, China}

\maketitle

\let\thefootnote\relax\footnotetext{This work was supported in part by the National Natural Science Foundation of China under Grant 61822109 and the Fok Ying Tung Education Foundation under Grant 161061.}
\begin{abstract}
	Anomaly detection in surveillance videos is a challenging task due to the diversity of anomalous video content and duration. In this paper, we consider video anomaly detection as a regression problem with respect to anomaly scores of video clips under weak supervision. Hence, we propose an anomaly detection framework, called Anomaly Regression Net (AR-Net), which only requires video-level labels in training stage. Further, to learn discriminative features for anomaly detection, we design a dynamic multiple-instance learning loss and a center loss for the proposed AR-Net. The former is used to enlarge the inter-class distance between anomalous and normal instances, while the latter is proposed to reduce the intra-class distance of normal instances. Comprehensive experiments are performed on a challenging benchmark: \textit{ShanghaiTech}. Our method yields a new state-of-the-art result for video anomaly detection on \textit{ShanghaiTech} dataset. 
\end{abstract}
\section{Introduction}
\label{sec:intro}
Video anomaly detection is an important yet challenging task in computer vision, and it is widely used in crime warning, intelligent video surveillance and evidence collection. According to the study~\cite{Zhong2019A}, there are two kinds of paradigms: unary classification and binary classification, for weakly-supervised video anomaly detection. Anomalies are usually defined as the video content patterns that are different from usual patterns in previous works~\cite{liu2018future}~\cite{ionescu2019object}~\cite{lu2013abnormal}~\cite{Luo2017remembering}~\cite{Hasan2016Learning}. Based on this definition, the unary classification paradigm-based methods only model usual patterns with normal training samples. However, it is impossible to collect all kinds of normal samples in a training set. Consequently, normal videos being different from training ones may tend to be false alarmed under this paradigm.

To address this issue, the binary classification paradigm was introduced, in which training data contains both anomalous and normal videos. Following the binary classification paradigm, some studies on anomaly detection ~\cite{Zhong2019A}~\cite{Sultani2018Real}~\cite{Yi2019Motion}~\cite{Zhang2019temporal} have been published. In~\cite{Zhong2019A}, video anomaly detection was formulated as a fully-supervised learning task under noise labels. As a correction, a Graph Convolutional Network (GCN) was proposed to train an action classifier. The GCN and the action classifier were optimized alternately.

\begin{figure}[t]	
	\begin{minipage}{0.45\linewidth}
		\centerline{\includegraphics{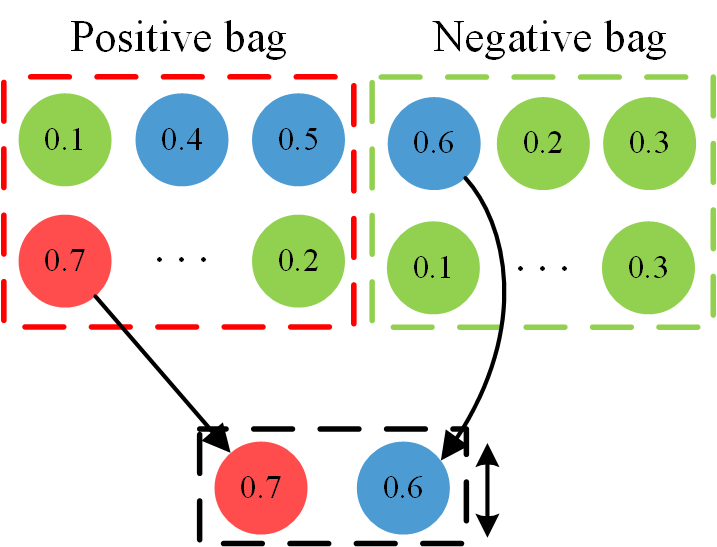}}	
		\centerline{(a)}	
	\end{minipage}
	\hfill	
	\begin{minipage}{0.45\linewidth}	
		\centerline{\includegraphics{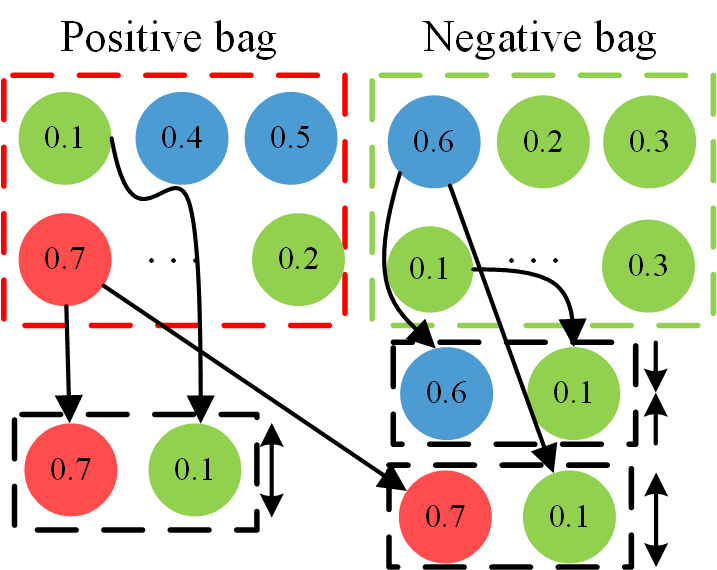}}	
		\centerline{(b)}	
	\end{minipage}
	\vfill	
	\begin{minipage}{0.45\linewidth}	
		\centerline{\includegraphics{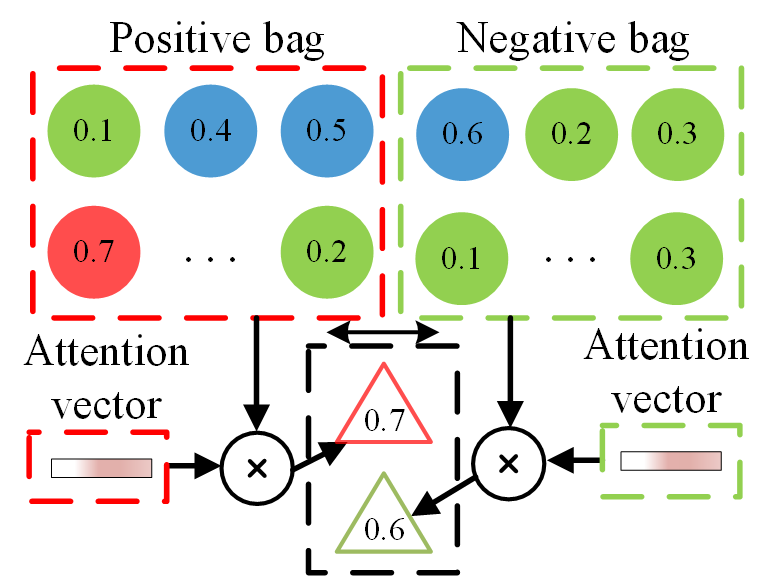}}	
		\centerline{(c)}	
	\end{minipage}
	\hfill	
	\begin{minipage}{0.45\linewidth}	
		\centerline{\includegraphics{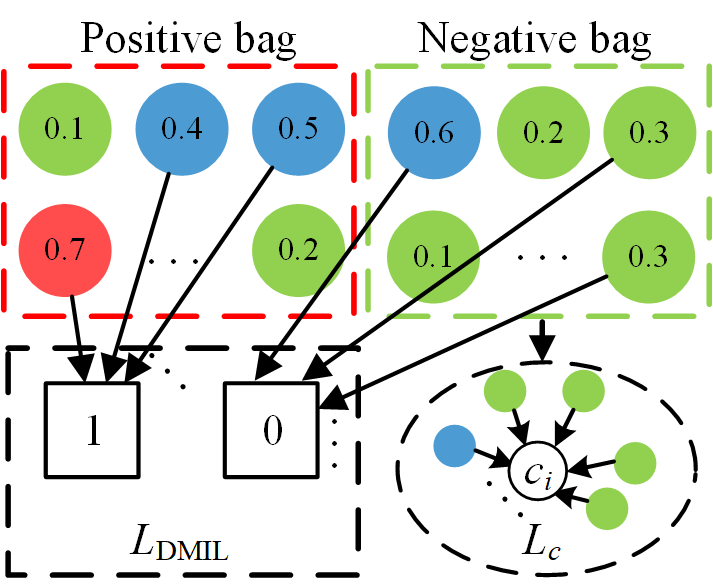}}	
		\centerline{(d)}	
	\end{minipage}	
	\caption{Comparison of loss functions for MIL-based anomaly detection methods. Float numbers in colored circles stand for anomaly scores of segments or clips, while those in colored triangles represent anomaly scores of videos. Binary numbers in black squares are video-level labels. (a) MIL ranking loss in~\cite{Sultani2018Real}. (b) Complementary inner bag loss in~\cite{Zhang2019temporal}. (c) Temporal ranking loss in~\cite{Yi2019Motion}. (d) Ours.}
	\label{fig:fig1}
\end{figure}

In this paper, we formulate video anomaly detection as a weakly-supervised learning problem following binary classification paradigm, where only video-level labels are involved in training stage. Recently, Multiple-Instance Learning (MIL) has become a major technique in several computer vision tasks including weakly-supervised temporal activity localization and classification~\cite{paul2018w}~\cite{narayan20193c} and weakly-supervised object detection~\cite{wan2019c}. There are also some MIL-based video anomaly detection studies~\cite{Sultani2018Real}~\cite{Yi2019Motion}~\cite{Zhang2019temporal}. Each training video is treated as a bag, and clips of videos are regarded as instances in these methods. An anomalous video is treated as a positive bag, and a normal one is presented as a negative bag. Sultani \textit{et al.}~\cite{Sultani2018Real} proposed a deep MIL ranking model with features extracted by C3D network~\cite{tran2015learning} as input. The deep MIL ranking loss was proposed for separating the anomaly scores of anomalous and normal instances. Zhang \textit{et al.}~\cite{Zhang2019temporal} proposed a complementary inner bag loss to reduce intra-class distances and enlarge inter-class distances of instances simultaneously. The highest and lowest anomaly scores of anomalous and normal videos were required. Zhu and Newsam~\cite{Yi2019Motion} proposed a temporal ranking loss calculated according to anomaly scores. The anomaly score of a video is the weighted sum of a video anomaly score vector and an attention vector. However, as shown in Fig~\ref{fig:fig1}, these methods adopted pair-wisely calculated losses, based on which the detection ability of models partly depend on batch size. In other words, the detection performance is partly limited by graphics memory. In this work, we look into a method that maximizes the inter-class distances and minimizes intra-class distances without instances from pair videos. 

We propose a framework, termed Anomaly Regression Network (AR-Net), and two novel losses to learn discriminative features under video-level weak supervision. As illustrated in Fig~\ref{fig:fig1}-(d), the dynamic multiple-instance learning loss ($L_{\text{DMIL}}$) is proposed to make features more separable. $L_{\text{DMIL}}$ is acquired by calculating the cross entropy between the anomaly scores of video clips and their corresponding video labels. The center loss is designed as the distances between anomaly scores of video clips and their corresponding average anomaly score $c_{i}$ in each training normal video. By minimizing the two losses, a discriminative feature representation can be obtained for video anomaly detection.

Comprehensive experiments are conducted on a benchmark: ShanghaiTech~\cite{luo2017revisit}. Our approach yields a new state-of-the-art result and obtains an absolute gain of 4.94\% in terms of Area Under the Curve (AUC) on ShanghaiTech dataset.
\begin{figure*}[ht]
	\centerline{\includegraphics[scale=0.9]{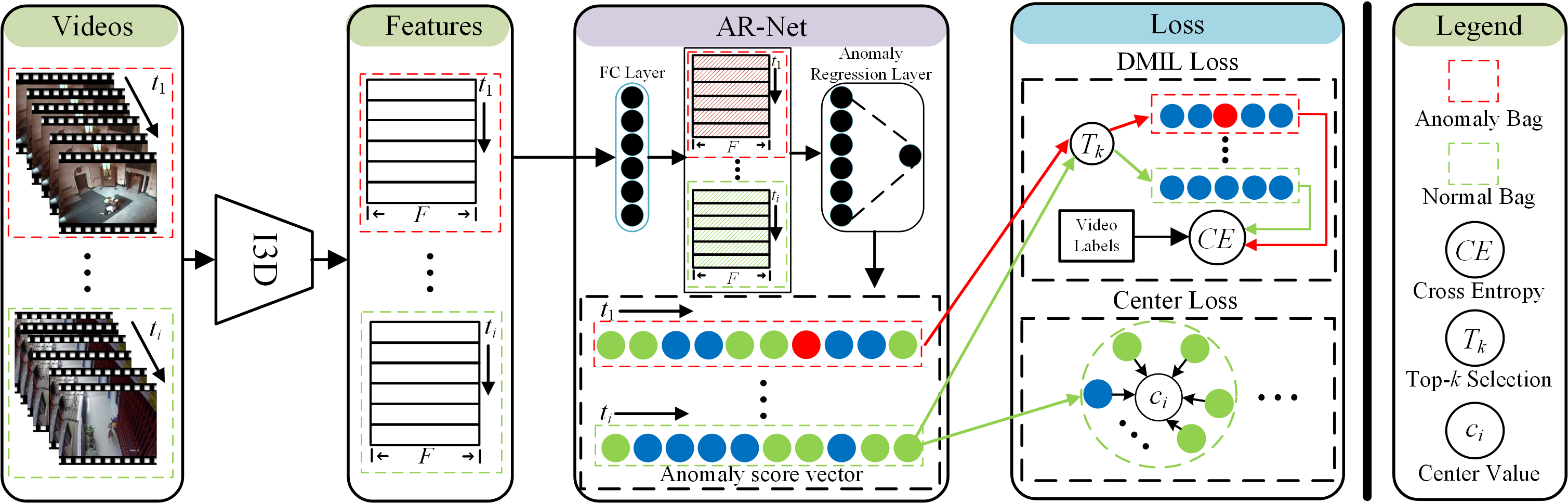}}
	\caption{Our model (AR-Net) with two proposed loss terms (DMIL, center), feature extractor and legend.}
	\label{fig:fig2}
\end{figure*}

\section{Proposed Method}

In this section, we first define the notations and problem statement. Then we describe the proposed feature extraction network. Finally, we present our AR-Net followed by a detailed description of the proposed losses.

\noindent\textbf{Problem Statement}: A training set consisting of \textit{n} videos is denoted by $\chi=\{\bm{x}_{i}\}_{i=1}^{n}$ in an anomaly detection dataset. The temporal duration of the dataset is defined as $\bm{T}=\{t_{i}\}_{i=1}^n$, where $t_i$ is the clip number of the $i$-th video. The video anomaly label set is denoted as $\bm{Y}=\{y_{i}\}_{i=1}^{n}$, where $y_i=\{0, 1\}$. In the testing stage, the predicted anomaly score vector of a video $\bm{x}$ is denoted as $\bm{s}=\{s^{j}\}_{j=1}^{t}$, where $s^j\in[0, 1]$, and $s^j$ is anomaly score of the $j$-th video clip.

\subsection{Feature Extraction}
To make use of both the appearance and motion information of the videos, Inflated 3D (I3D)~\cite{carreira2017quo}, pretrained on the Kinetics~\cite{carreira2017quo} dataset, is used as the feature extraction network. An input video is divided into non-overlapped clips, each of which contains 16 consecutive frames. The RGB and Optical-Flow versions of I3D are denoted by I3D$^{\text{RGB}}$ and I3D$^{\text{Optical-Flow}}$ respectively. The former takes RGB frames as input while the latter takes Optical-Flow frames. We concatenate features from penultimate layer of the I3D$^{\text{RGB}}$ and I3D$^{\text{Optical-Flow}}$ as our final feature representation of video clips.

The feature matrix $\bm{X}_{i}$ shown in Fig~\ref{fig:fig2} is composed of features from the training video $\bm{x}_{i}$.The dimension of $\bm{X}_{i}$ is $F\times t_{i}$, where $F$ is the dimension of clip features. $\bm{X}_{i}$ rather than raw video clips is fed to our AR-Net.

\subsection{Anomaly Regression Network}

The architecture of AR-Net is shown in Fig~\ref{fig:fig2}. The Fully Connected Layer (FC-Layer) and Anomaly Regression Layer (AR-Layer) in AR-Net require only video-level labels for video anomaly detection. We adopt ReLU~\cite{nair2010rectified} as the activation function of FC-Layer. To avoid overfitting, Dropout~\cite{srivastava2014dropout} is introduced to FC-Layer and can be formalized as follows:

\begin{eqnarray}
	\label{eqn:eqn1}
	\bm{X}^{\text{FC}}_{i}=D(max(0,\bm{W}_{\text{FC}}{\bm{X}_{i}}+\bm{b}_{\text{FC}}))
\end{eqnarray}
where $D(\cdot)$ denotes Dropout, $\bm{W}_{\text{FC}}\in \mathbb{R}^{F \times F}$ and $\bm{b}_{\text{FC}}\in \mathbb{R}^{F \times 1}$ are learnable parameters to be optimized from the training data, and $\bm{X}_i^{\text{FC}}\in \mathbb{R}^{F \times t_i}$ is the $i$-th video feature output from output features of the FC-Layer.

We establish a mapping function between the representations $\bm{X}_i^{\text{FC}}$ and anomaly score vectors $\bm{s}_{i}$ by AR-Layer, which is a fully connected layer. The AR-Layer can be represented as follows:
\begin{eqnarray}
	\label{eqn:eqn2}
	\bm{s}_{i}=\frac{1}{1+exp(\bm{W}_{\text{AR}}{\bm{X}^{\text{FC}}_{i}}+b_{\text{AR}})}
\end{eqnarray}
where $\bm{W}_{\text{AR}}\in \mathbb{R}^{1 \times F}$, ${b}_{\text{AR}}\in \mathbb{R}^{1}$ are learnable parameters and $\bm{s}_{i}\in \mathbb{R}^{1\times t_{i}}$. The anomaly score vectors $\bm{s}_{i}$ represent the probabilities that instances are classified as anomalies.

\subsection{Dynamic Multiple-Instance Learning Loss}
As discussed in Section~\ref{sec:intro}, the video anomaly detection is treated as a MIL task in this paper. In MIL, a positive bag contains at least one positive instance and a negative bag contains no positive instances, \textit{i.e}., an abnormal video contains at least one anomalous event and a normal video contains no anomalous events. To enlarge the inter-class distance between anomalous and normal instances under a weak supervision, inspired by the $k$-max MIL loss in~\cite{paul2018w}~\cite{narayan20193c}, we propose a Dynamic Multiple-Instance learning (DMIL) loss that takes the diversity of video duration into consideration.

Different from the max-selection method involved in MIL-based loss function used in~\cite{Sultani2018Real}~\cite{Yi2019Motion}~\cite{Zhang2019temporal}, we introduce $k$-max selection method, which is used in~\cite{paul2018w}~\cite{narayan20193c}, to obtain the $k$-max anomaly scores. The $k$ is determined based on the number of clips in a video. Specifically,

\begin{eqnarray}
	\label{eqn:eqn3}
	k_{i}= \lceil {\frac{t_{i}}{\alpha}} \rceil
\end{eqnarray}
where $\alpha$ is a hyperparameter. Thus, the $k$-max anomaly scores of the $i$-th video can be represented as,

%

\begin{eqnarray}
	\label{eqn:eqn4_3}
	\left\{\begin{aligned}
		&\bm{p}_{i} = sort(\bm{s}_{i}) \\ 
		&\bm{S}_{i} = \{p_i^{j}\left|\right. j=1,2,...,k_i\}
	\end{aligned}\right.
\end{eqnarray}
where $\bm{s}_{i}$ is anomaly score vector of the $i$-th video, $sort(\cdot)$ is a descending sort operator and $\bm{p}_{i}$ is the sorted $\bm{s}_{i}$. Thus, $\bm{S}_{i}$ consists of top-$k_{i}$ elements in $\bm{s}_{i}$. The DMIL loss can then be represented as follows:

\begin{eqnarray}
	\label{eqn:eqn5}
	L_{\text{DMIL}}=\frac{1}{k_{i}} \sum_{s_{i}^{j}\in{\bm{S}_{i}}} [-y_{i}log(s_{i}^{j})\nonumber \\
	+ (1-y_{i})log(1-s_{i}^{j})]
\end{eqnarray}
where $y_{i}=\{0, 1\}$ is the video anomaly label. Furthermore, instead of calculating the cross-entropy between the average of selected $k$ scores and the video label in~\cite{paul2018w}~\cite{narayan20193c}, we calculate the cross-entropy between each of the selected $k$ scores and the video label as the instance loss respectively. Noise labels will affect anomaly scores of the sample features from which an average anomaly score is calculated. While our DMIL loss focuses on individual anomaly scores rather than an average one. Thus, this loss keeps errors brought by noise labels from propagating.

\subsection{Center loss for Anomaly Scores Regression}

The objective of the DMIL loss is to enlarge the inter-class distance of instances. However, both the max and $k$-max selection method inevitably produce wrong label assignment, since the anomaly scores of normal clips and abnormal clips in the abnormal video are similar in early training stage. As a result, the intra-class distance of normal instances is unfortunately enlarged by the DMIL loss, and this will reduce detection accuracy in testing stage. 

Inspired by the center loss in~\cite{wen2016centerloss}, we propose a novel center loss for anomaly score regression to address the above-mentioned issue. In~\cite{wen2016centerloss}, the center loss learns the feature center of each class and penalizes the distance between the feature representations and their corresponding class centers. In our case, the center loss proposed for anomaly score regression gathers the anomaly scores of normal video clips.

Our center loss for anomaly score regression can be represented as,

\begin{eqnarray}
	\label{eqn:eqn6}
	L_{c}=\left\{\begin{aligned}
		& \frac{1}{t_{i}}\sum_{j=1}^{t_{i}}\begin{Vmatrix}
			s_{i}^{j}-c_{i}
		\end{Vmatrix}^{2}_{2},& \text{if  }y_{i}=0\\ 
		& 0,& \text{otherwise}
	\end{aligned}\right.
\end{eqnarray}
\begin{eqnarray}
	\label{eqn:eqn7}
	c_{i}=\frac{1}{t_{i}}\sum_{j=1}^{t_{i}}s_{i}^{j}
\end{eqnarray}
where $c_{i}$ is the center of anomaly score vector $\bm{s}_{i}$ of the $i$-th video.

\subsection{Optimization}
The total loss function of the AR-Net can be represented as follows:
\begin{eqnarray}
	\label{eqn:eqn8}
	L=L_{\text{DMIL}}+\lambda L_{c}
\end{eqnarray}
To achieve a balance between the two losses in training stage, we empirically set $\lambda$ = 20.

\section{Experiments}

\subsection{Experiments Setup}

\textbf{Datasets}: The proposed AR-Net is evaluated on a challenging dataset containing untrimmed videos with variable scenes, contents and durations. ShanghaiTech~\cite{luo2017revisit} is a dataset that contains 437 videos with 130 anomalies on 13 scenes. However, this dataset is proposed for unary-classification, so all the training videos are normal~\cite{luo2017revisit}. To make binary-classification available, we adopt the split version proposed in~\cite{Zhong2019A}. Specifically, there are 238 training videos and 199 testing videos.

\noindent\textbf{Evaluation Metric}: Similar to previous works~\cite{Zhong2019A}~\cite{luo2017revisit}~\cite{Sultani2018Real}, we use an Area Under of Curve (AUC) of the frame-level Receiver Operating Characteristics (ROC) and False Alarm Rate (FAR) with threshold 0.5 as the evaluation metrics. In the video anomaly detection task, the higher AUC demonstrates, the better the model performs, and lower FAR on a normal video implies stronger robustness of an anomaly detection method.

\noindent\textbf{Implementation Details}: We combine I3D$^{\text{RGB}}$ and I3D$^{\text{Optical-Flow}}$ as our feature extractor, denoted as I3D$^{\text{Conc}}$. The feature of I3D$^{\text{Conc}}$ is the concatenation of features from I3D$^{\text{RGB}}$ and I3D$^{\text{Optical-Flow}}$. Besides, our feature extractor is not fine-tuned. The Optical-Flow frames of each clip are generated based on TV-L1 algorithm~\cite{TV-L1}. Empirically, we set $\alpha = 4$ for ShanghaiTech dataset. The weights of the AR-Net are initialized by Xavier method~\cite{glorot2010understanding}, and the Dropout probability for FC-Layer is 0.7. We adopt the Adam optimizer~\cite{kingma2014adam} with a batch size of 60, in which 30 normal videos and 30 abnormal ones are randomly selected from the training set. The learning rate is always $10^{-4}$ in our experiments.

\begin{table}[ht]\footnotesize
	\caption{AUC and FAR of the proposed method against 3 existing methods. The $^{\S}$, $^{\dagger}$ and $^{\ddagger}$ indicate the anomaly detection model based on C3D, TSN$^{\text{RGB}}$ and TSN$^{\text{Optical-Flow}}$ in~\cite{Zhong2019A}, respectively.}
	\centering{\begin{tabular}{@{}lllll@{}} \hline
			\multirow{2}{*}{Methods}& \multicolumn{2}{c}{ShanghaiTech}\\
			& AUC(\%)& FAR(\%)\\ \hline 
			Sultani \textit{et al.}~\cite{Sultani2018Real}& 86.30& 0.15\\
			Zhang \textit{et al.}~\cite{Zhang2019temporal}& 82.50& 0.10\\
			Zhong $\textit{et al.}^{\S}$~\cite{Zhong2019A}& 76.44& $-$\\
			Zhong $\textit{et al.}^{\dagger}$~\cite{Zhong2019A}& 84.44& $-$\\
			Zhong $\textit{et al.}^{\ddagger}$~\cite{Zhong2019A}& 84.13& $-$\\
			AR-Net& \textbf{91.24}& \textbf{0.10}\\ \hline
		\end{tabular}} \label{table:tb1}
	\end{table}
	\subsection{Comparison Results}
	
	Table~\ref{table:tb1} shows the comparison of our method against existing approaches~\cite{Sultani2018Real}~\cite{Zhong2019A}~\cite{Zhang2019temporal} on the ShanghaiTech.
		
	In order to present a comparison against MIL-based works on ShanghaiTech, we reproduced the method in ~\cite{Zhang2019temporal} and adopted the open source code provided by Sultani \textit{et al.}~\cite{Sultani2018Real} to conduct the anomaly detection. The above two models are obtained by pretrained C3D. As shown in Table~\ref{table:tb1}, ~\cite{Zhang2019temporal} achieves a frame-level AUC of 82.50\%. Meanwhile,~\cite{Sultani2018Real} performs a frame-level AUC of 86.30\%, outperforming the best existing method~\cite{Zhong2019A}. Our method substantially exceeds both Sultani \textit{et al.}~\cite{Sultani2018Real} and Zhong $\textit{et al.}^{\dagger}$~\cite{Zhong2019A} with a frame-level AUC of 91.24\%. Furthermore, our approach is the only one surpassing 90\% in terms of AUC on ShanghaiTech.

	\begin{table}[h]\footnotesize
		\caption{AUC and FAR of different loss functions.}
		\centering{\begin{tabular}{@{}lllll@{}} \hline
				\multirow{2}{*}{Losses}& \multicolumn{2}{c}{ShanghaiTech}\\
			    & AUC(\%)& FAR(\%)\\ \hline
				Baseline:$L_{\text{$k$-max MIL}}$& 86.50& 0.93\\ 
				Ours:$L_{\text{DMIL}}$& 89.10& 0.21\\
				Ours:$L_{\text{DMIL}}+L_{c}$& 91.24& 0.10\\ \hline
			\end{tabular}} \label{table:tb2}
		\end{table}
		
		\begin{table}[h]\footnotesize
			\caption{AUC and FAR of different feature extractors.}
			\centering{\begin{tabular}{@{}lllll@{}} \hline
					\multirow{2}{*}{Feature extractor}& \multicolumn{2}{c}{ShanghaiTech}\\
					& AUC(\%)& FAR(\%)\\ \hline
					I3D$^{\text{RGB}}$& 85.38& 0.27\\ 
					I3D$^{\text{Optical-Flow}}$& 82.34& 0.37\\
					I3D$^{\text{Conc}}$& 91.24& 0.10\\ \hline
				\end{tabular}} \label{table:tb3}
			\end{table}
		
		
%
		\subsection{Ablation Study}
		
		The comparison results by using different loss functions in Table~\ref{table:tb2} illustrate the boost brought by the proposed $L_{c}$ and $L_{\text{DMIL}}$ in our AR-Net. AR-Net that involves k-max selection-based MIL loss ($L_{\text{$k$-max MIL}}$) is treated as the baseline in our ablation study. It achieves a frame-level AUC of 86.50\% on ShanghaiTech. While the proposed DMIL loss-based AR-Net boosts the performance by obtaining a frame-level AUC of 89.10\% on ShanghaiTech. Besides, with the help of the proposed $L_{c}$, the FAR on ShanghaiTech is reduced to 1/9 of those achieved by baseline.
		
		
		To demonstrate the performance brought by video appearance and motion information, we compare the anomaly detection results based on different feature extractors. As show in Table~\ref{table:tb3}, AR-Net with I3D$^{\text{RGB}}$ achieves a frame-level AUC of 85.38\%. And the I3D$^{\text{Optical-Flow}}$ based AR-Net achieves a frame-level AUC of 82.34\%. The AR-Net with I3D$^{\text{Conc}}$ boosts the performance with a frame-level AUC of 91.24\%.
		\begin{figure}[t]
			\begin{minipage}{0.40\linewidth}
				\centerline{\includegraphics{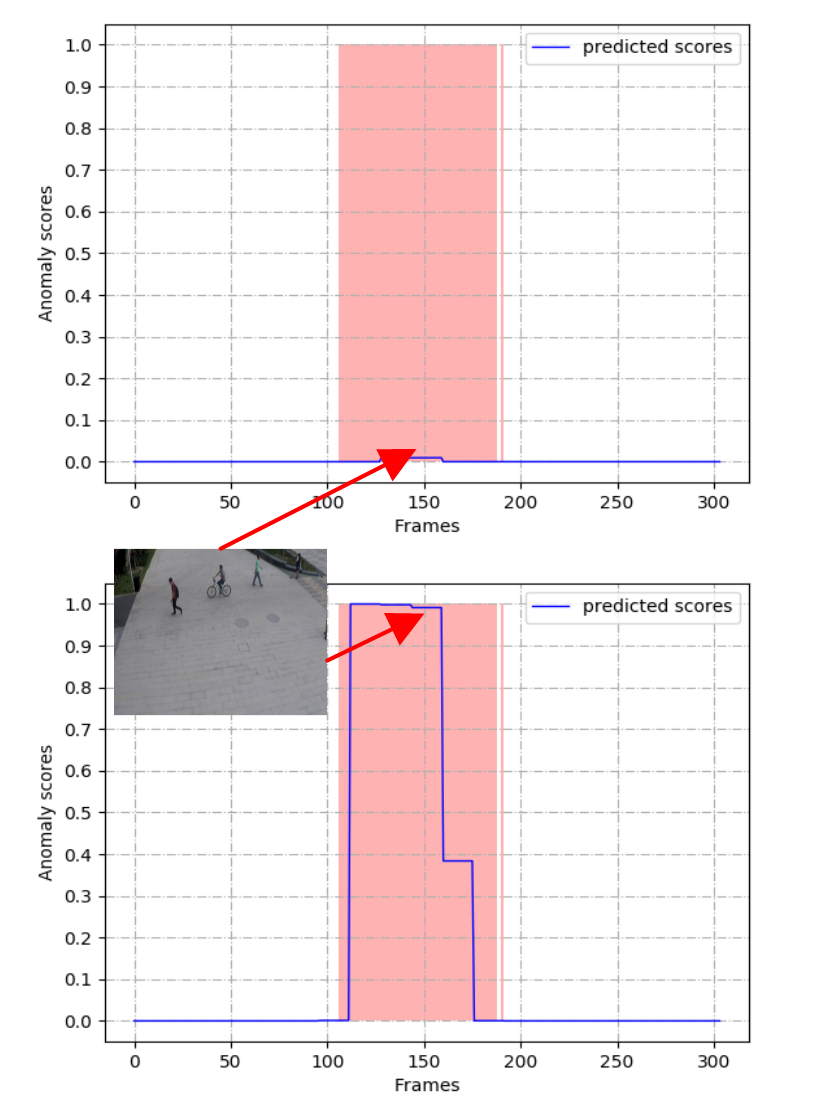}}	
				\centerline{(a) 01\_0177}	
			\end{minipage}
			\hfill	
			\begin{minipage}{0.4\linewidth}
				\centerline{\includegraphics{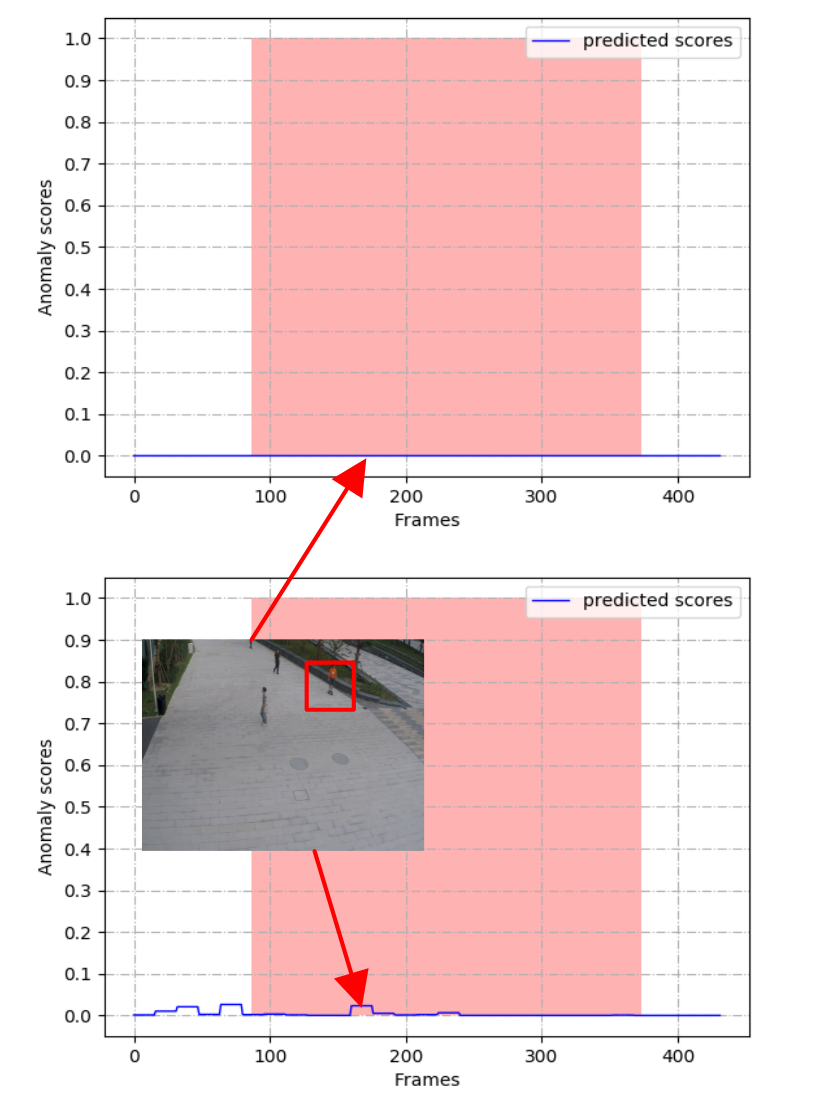}}	
				\centerline{(b) 01\_0015}	
			\end{minipage}
			\caption{Comparison visualization of testing results between~\cite{Sultani2018Real} and ours.}
			\label{fig:fig4}
		\end{figure}
		\begin{figure}[t]
			\centering
			\includegraphics[width=0.50\textwidth]{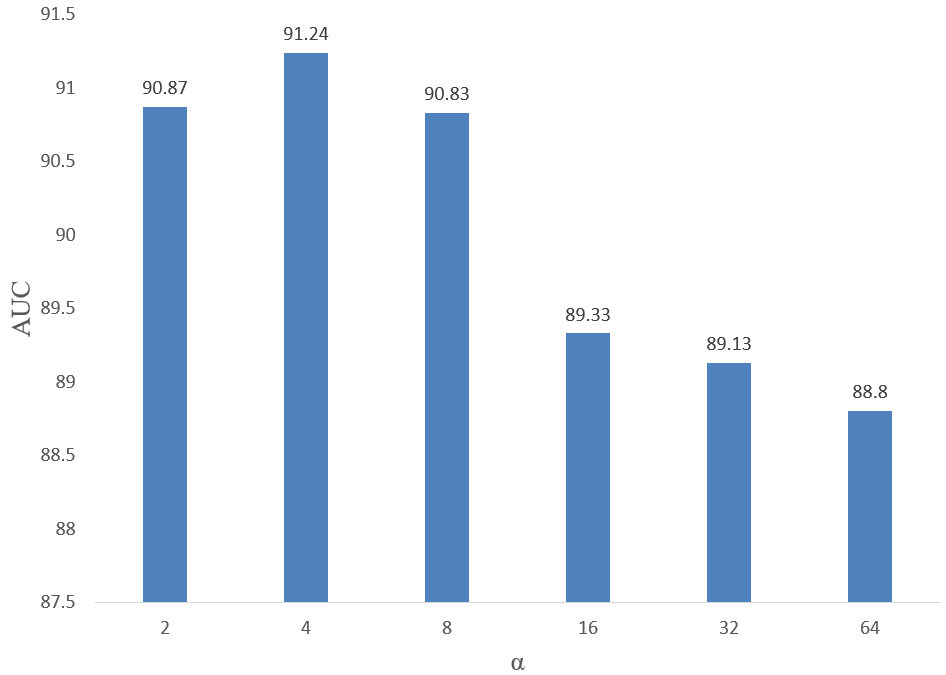}	
			\caption[width=0.4\textwidth]{AUC of different $\alpha$ values.}
			\label{fig:fig3}
		\end{figure}	
		
		\subsection{Qualitative Analysis}
		
		As shown in Table~\ref{table:tb1}, our method surpasses most of the recent MIL-based works. In ShanghaiTech, sometimes anomaly makes up a tiny part of a whole video. In segmented-based methods~\cite{Sultani2018Real}~\cite{Zhang2019temporal}, although each video is divided into 32 non-overlapped segments, the anomalous events in each segment may still account for little parts. In other words, the features of anomalous frames to be overwhelmed by normal ones in a segment. As a result, segments containing anomaly tend to be regarded as normal patterns, i.e., the anomaly detection model trained by these segments lacks capability to detect short-term anomaly. While in clip-based methods like~\cite{paul2018w}~\cite{narayan20193c} and ours, videos are divided into clips with fixed number of frames. Anomaly detection models using this strategy avoids anomaly being overwhelmed by normal frames and are able to recognize short-term anomaly. As shown in Fig.3-(a), the model in~\cite{Sultani2018Real} (illustrated in the 1st row) does not recognize illegal bicycling while ours (in the 2nd row) does.
		
		As shown in the 1st row of Fig.3-(b), both of [7] (illustrated in the 1st row) and our method (in the 2nd row) fail to detect abnormal events in `01\_0015'. The reasons are: 1) the anomaly, marked with a red box, only takes up a local part of the video scene, while methods with global input are prone to ignoring local anomalies 2) the anomaly of skateboarding on the sidewalk are not visually distinguishable from normal behaviors, hence the anomalous clips and non-ones are not separable. In conclusion, video anomaly detection in these kinds of scenes is still a big challenge for current models.
		
		In order to gain insight into the hyperparameter $\alpha$, we perform experiments using the I3D$^{\text{Conc}}$ feature extractor with different values of $\alpha$, as shown in Fig.4. In fact, the $\alpha$ determines the proportion of noisy-label instances in training stage. Although higher $\alpha$ results in a smaller proportion of noisy-label instances, some anomalous instances will be ignored in training stage. This leads to insufficient diversity of training anomalous instances, which reduces the frame-level AUC of the AR-Net. There is at most 2.44\% decrease on frame-level AUC when $\alpha$ is set to 8, 16, 32, or 64. Moreover, the lower $\alpha$ causes more normal instances being labeled as the anomalous ones in training stage. This also reduces the frame-level AUC of the AR-Net. Fig 4 shows that AR-Net suffers 0.37\% decrease on frame-level AUC when $\alpha$ is lower than 4.

		\section{Conclusion}
		
		In this paper, we propose a MIL-based anomaly regression network for video anomaly detection. Besides, we design a dynamic loss $L_{\text{DMIL}}$ to learn separable features and a center loss $L_{c}$ to correct the anomaly scores output by our AR-Net. By optimizing the parameters of AR-Net under weak supervision, the dynamic multiple-instance learning loss avoids false alarm caused by interference between clip features. While the center regression loss suppresses label noise by smoothing the distribution of anomaly scores. In addition, the clip-based instance generation strategy benefits to short-term anomaly detection. Experiments on a challenging datasets clearly demonstrate the effectiveness of our approach for video anomaly detection. In the future, we will investigate to model temporal relation between instances to obtain more robustness.

		\bibliographystyle{IEEEbib}
		\bibliography{wanbib_short}

\begin{thebibliography}{10}

\bibitem{Zhong2019A}
Jia-Xing Zhong, Nannan Li, Weijie Kong, Shan Liu, Thomas~H. Li, and Ge~Li,
\newblock ``Graph convolutional label noise cleaner: Train a plug-and-play
  action classifier for anomaly detection,''
\newblock in {\em IEEE Conference on Computer Vision and Pattern Recognition},
  2019, pp. 1237--1246.

\bibitem{liu2018future}
Wen Liu, Weixin Luo, Dongze Lian, and Shenghua Gao,
\newblock ``Future frame prediction for anomaly detection--a new baseline,''
\newblock in {\em IEEE Conference on Computer Vision and Pattern Recognition},
  2018, pp. 6536--6545.

\bibitem{ionescu2019object}
Radu~Tudor Ionescu, Fahad~Shahbaz Khan, Mariana-Iuliana Georgescu, and Ling
  Shao,
\newblock ``Object-centric auto-encoders and dummy anomalies for abnormal event
  detection in video,''
\newblock in {\em IEEE Conference on Computer Vision and Pattern Recognition},
  2019, pp. 7842--7851.

\bibitem{lu2013abnormal}
Cewu Lu, Jianping Shi, and Jiaya Jia,
\newblock ``Abnormal event detection at 150 fps in matlab,''
\newblock in {\em IEEE International Conference on Computer Vision}, 2013, pp.
  2720--2727.

\bibitem{Luo2017remembering}
Weixin Luo, Wen Liu, and Shenghua Gao,
\newblock ``Remembering history with convolutional lstm for anomaly
  detection,''
\newblock in {\em IEEE International Conference on Multimedia and Expo}, 2017,
  pp. 439--444.

\bibitem{Hasan2016Learning}
Mahmudul Hasan, Jonghyun Choi, Jan Neumann, Amit~K. Roy-Chowdhury, and Larry~S.
  Davis,
\newblock ``Learning temporal regularity in video sequences,''
\newblock in {\em IEEE Conference on Computer Vision and Pattern Recognition},
  2016, pp. 733--742.

\bibitem{Sultani2018Real}
Waqas Sultani, Chen Chen, and Mubarak Shah,
\newblock ``Real-world anomaly detection in surveillance videos,''
\newblock in {\em IEEE Conference on Computer Vision and Pattern Recognition},
  2018, pp. 6479--6488.

\bibitem{Yi2019Motion}
Yi~Zhu and Shawn~D. Newsam,
\newblock ``Motion-aware feature for improved video anomaly detection,''
\newblock {\em CoRR}, vol. abs/1907.10211, 2019.

\bibitem{Zhang2019temporal}
Jiangong Zhang, Laiyun Qing, and Jun Miao,
\newblock ``Temporal convolutional network with complementary inner bag loss
  for weakly supervised anomaly detection,''
\newblock in {\em IEEE International Conference on Image Processing}, 2019, pp.
  4030--4034.

\bibitem{paul2018w}
Sujoy Paul, Sourya Roy, and Amit~K Roy-Chowdhury,
\newblock ``{W-TALC}: Weakly-supervised temporal activity localization and
  classification,''
\newblock in {\em European Conference on Computer Vision}, 2018, pp. 563--579.

\bibitem{narayan20193c}
Sanath Narayan, Hisham Cholakkal, Fahad~Shahbaz Khan, and Ling Shao,
\newblock ``3{C}-{N}et: Category count and center loss for weakly-supervised
  action localization,''
\newblock in {\em IEEE International Conference on Computer Vision}, 2019, pp.
  8679--8687.

\bibitem{wan2019c}
Fang Wan, Chang Liu, Wei Ke, Xiangyang Ji, Jianbin Jiao, and Qixiang Ye,
\newblock ``{C-MIL}: Continuation multiple instance learning for weakly
  supervised object detection,''
\newblock in {\em IEEE Conference on Computer Vision and Pattern Recognition},
  2019, pp. 2199--2208.

\bibitem{tran2015learning}
Du~Tran, Lubomir Bourdev, Rob Fergus, Lorenzo Torresani, and Manohar Paluri,
\newblock ``Learning spatiotemporal features with 3d convolutional networks,''
\newblock in {\em IEEE Conference on Computer Vision and Pattern Recognition},
  2015, pp. 4489--4497.

\bibitem{luo2017revisit}
Weixin Luo, Wen Liu, and Shenghua Gao,
\newblock ``A revisit of sparse coding based anomaly detection in stacked rnn
  framework,''
\newblock in {\em IEEE International Conference on Computer Vision}, 2017, pp.
  341--349.

\bibitem{carreira2017quo}
Joao Carreira and Andrew Zisserman,
\newblock ``Quo vadis, action recognition? a new model and the kinetics
  dataset,''
\newblock in {\em IEEE Conference on Computer Vision and Pattern Recognition},
  2017, pp. 6299--6308.

\bibitem{nair2010rectified}
Vinod Nair and Geoffrey~E Hinton,
\newblock ``Rectified linear units improve restricted boltzmann machines,''
\newblock in {\em International Conference on Machine Learning}, 2010, pp.
  807--814.

\bibitem{srivastava2014dropout}
Nitish Srivastava, Geoffrey Hinton, Alex Krizhevsky, Ilya Sutskever, and Ruslan
  Salakhutdinov,
\newblock ``Dropout: a simple way to prevent neural networks from
  overfitting,''
\newblock {\em The Journal of Machine Learning Research}, vol. 15, no. 1, pp.
  1929--1958, 2014.

\bibitem{wen2016centerloss}
Yandong Wen, Kaipeng Zhang, Zhifeng Li, and Yu~Qiao,
\newblock ``A discriminative feature learning approach for deep face
  recognition,''
\newblock in {\em European Conference on Computer Vision}, 2016, pp. 499--515.

\bibitem{TV-L1}
Frank Steinbrücker, Thomas Pock, and Daniel Cremers,
\newblock ``Large displacement optical flow computation withoutwarping,''
\newblock in {\em IEEE International Conference on Computer Vision}, 2009, pp.
  1609--1614.

\bibitem{glorot2010understanding}
Xavier Glorot and Yoshua Bengio,
\newblock ``Understanding the difficulty of training deep feedforward neural
  networks,''
\newblock in {\em Proceedings of the International Conference on Artificial
  Intelligence and Statistics}, 2010, pp. 249--256.

\bibitem{kingma2014adam}
Diederik~P. Kingma and Jimmy Ba,
\newblock ``Adam: {A} method for stochastic optimization,''
\newblock in {\em International Conference on Learning Representations}, 2015.

\end{thebibliography}
		
	\end{document}